\documentclass[11pt]{article}

\usepackage[final]{acl}

\usepackage{times}
\usepackage{latexsym}

\usepackage[T1]{fontenc}
\usepackage[utf8]{inputenc}

\usepackage{microtype}

\usepackage{inconsolata}

\usepackage{graphicx}

\usepackage{hyperref}
\usepackage{url}
\usepackage{graphicx}
\usepackage{wrapfig}
\usepackage[most]{tcolorbox}
\usepackage{enumitem}
\usepackage{booktabs}
\usepackage{multirow}
\usepackage{amsfonts} % This line loads the amsfonts package
\usepackage{amsmath} % Often used with amsfonts for enhanced math typesetting
\usepackage{mathtools}

\newtcolorbox{findingbox}[1][]{%
  enhanced,
  colback=gray!10,            % 正文灰色再深一点
  colframe=blue!60!black,     % 边框颜色与标题栏协调
  colbacktitle=blue!50!black, % 标题栏蓝灰色
  coltitle=white,             % 标题白字
  fonttitle=\bfseries\large,  % 标题稍大加粗
  boxrule=0.8pt,
  title=#1,
  before skip=10pt,
  after skip=10pt,
  left=8pt,right=8pt,top=6pt,bottom=6pt
}

\title{Turn-PPO: Turn-Level Advantage Estimation with PPO for Improved Multi-Turn RL in Agentic LLMs}

\author{
Junbo Li$^{1}$\thanks{Work done during internship at Amazon.} \And
Peng Zhou$^{2}$ \And
Rui Meng$^{2}$ \And
Meet P. Vadera$^{2}$ \And
Lihong Li$^{2}$ \And
Yang Li$^{2}$ \AND
\\[-3em]
$^{1}$The University of Texas at Austin \qquad
$^{2}$Amazon
}

\begin{document}

\maketitle

\begin{abstract}
    Reinforcement learning (RL) has re-emerged as a natural approach for training interactive LLM agents in real-world environments. However, directly applying the widely used Group Relative Policy Optimization (GRPO) algorithm to multi-turn tasks exposes notable limitations, particularly in scenarios requiring long-horizon reasoning. To address these challenges, we investigate more stable and effective advantage estimation strategies, especially for multi-turn settings.
    We first explore Proximal Policy Optimization (PPO) as an alternative and find it to be more robust than GRPO. To further enhance PPO in multi-turn scenarios, we introduce turn-PPO, a variant that operates on a turn-level MDP formulation, as opposed to the commonly used token-level MDP. Our results on the WebShop and Sokoban datasets demonstrate the effectiveness of turn-PPO, both with and without long reasoning components. 
\end{abstract}

\section{Introduction}
LLMs have shown strong potential for tool use and environment interaction \citep{wang2025ragen,feng2025group,wei2025webagent}. 
In these settings, reinforcement learning (RL) is commonly preferred over offline algorithms \citep{rafailov2023direct, ethayarajh2024kto, lipipa} because obtaining high-quality, step-by-step supervision for complex interactions is difficult. Instead, RL enables the model to optimize its policy directly from environmental feedback and sparse rewards, allowing it to learn effective strategies without optimal reference trajectories.

Unlike single-turn reasoning tasks, where all response tokens are generated in a single pass and the problem can be framed as a bandit setting, multi-turn tool-use scenarios are better modeled as Markov Decision Processes (MDPs), which involve sequential decision-making and state transitions. The current widely used RL method for this multi-turn setup is a direct adaptation of GRPO \citep{wang2025ragen,wei2025webagent}. This method samples multiple multi-turn trajectories per question, then estimates the advantage for all tokens in a trajectory by normalizing the trajectory-level reward using the group's mean and standard deviation. However, this adaptation faces several challenges in the multi-turn context that can impair optimization effectiveness: 
1) Environment interactions are not fully controllable, leading to \textbf{higher sampling variance} compared to single-turn settings, which makes advantage estimation less stable.
2) Different turns within a trajectory \textbf{contribute unequally} to the final reward, so applying the same advantage uniformly to all tokens across turns can introduce inaccuracies.
Several prior works \citep{feng2025group,zeng2025reinforcing} attempt to address this issue by incorporating turn-level advantage estimates. However, these methods are specialized for particular settings and can introduce additional biases, limiting the generalizability.

To address these issues, we reintroduce PPO \citep{schulman2017proximal} into multi-turn agentic LLM training. 
Unlike GRPO, which relies on multiple rollouts to estimate advantages and often suffers from instability, PPO leverages a learnable critic model for advantage estimation. 
Because it is designed for multi-step MDPs, PPO naturally supports effective training through Generalized Advantage Estimation (GAE). 
In existing multi-turn implementations, however, PPO is typically applied by treating each token as a separate MDP step, a practice inherited from single-turn settings. 
We show that this formulation is suboptimal for multi-turn tasks because it complicates critic learning due to the mismatch between token-level granularity and the underlying task structure. 
To improve PPO’s effectiveness, we propose \textbf{turn-PPO}, which redefines the MDP formulation at the turn level. 
Specifically, we treat the entire input and output of a turn as a single state–action pair rather than operating at the token level. 
This results in a more coherent representation, enables the critic to learn more accurately, and produces more reliable advantage estimates.

\paragraph{Contributions} Our main contributions are as follows:
\begin{itemize}
    \item We identify and analyze the instability of sampling-based advantage estimation used in GRPO and its variants in multi-turn settings.
    \item We show that PPO achieves greater training stability and efficiency than GRPO due to its learnable critic and more accurate advantage estimation.
    \item We introduce \textbf{turn-PPO}, a variant of token-PPO, and demonstrate its superior performance across diverse tasks and settings.
\end{itemize}

\subsection{Related work}
\paragraph{Multi-turn agentic LLMs}
It is well established that, rather than solving a problem in a single step, LLMs can reason and act step by step with the help of external tools \citep{yao2023react} and use intermediate results to guide future reasoning steps. When augmented with tool-calling abilities, they can interact with external environments to solve a wide range of tasks, including web navigation \citep{wei2025webagent, yao2022webshop, gur2023real, putta2024agent, jin2025search, qi2024webrl}, GUI automation \citep{qin2025ui, wang2025ui}, and embodied AI control \citep{li2024embodied, huang2022language}. In this work, we restrict our study to text-only environments, enabling us to concentrate purely on improving reinforcement learning algorithms for training LLMs.

\paragraph{RL for LLMs}
By modeling LLMs as MDPs, we can directly apply policy-based reinforcement learning algorithms to train them. In the single-turn setting, strictly on-policy REINFORCE-style methods have been explored, including Reinforce++ \citep{hu2025reinforce++}, RLOO \citep{ahmadian2024back}, and ReMax \citep{li2023remax}, as well as off-policy PPO-style methods such as VinePPO \citep{kazemnejad2024vineppo}, VC-PPO \citep{yuan2025s}, VAPO \citep{yue2025vapo}. Building on PPO, DeepSeek-R1 \citep{guo2025deepseek} introduces GRPO \citep{shao2024deepseekmath}, which replaces the learned critic with sample-based advantage estimation. Several follow-up works, including DAPO \citep{yu2025dapo}, Dr. GRPO \citep{liu2025understanding}, Reinforce-Rej \citep{xiong2025minimalist}, and GSPO \citep{zheng2025group} further refine or extend GRPO to improve stability and performance.

Extending from the single-turn to the multi-turn setting, several RL variants have been proposed. ArCHer \citep{zhou2024archer} introduces a hierarchical RL framework that employs an actor–critic algorithm at the turn level and a policy-gradient update at the token level. \cite{yin2025segmenting} trained segment-level reward model to obtain dense supervision for PPO. WebAgent-R1 \citep{wei2025webagent} introduces M-GRPO, a direct adaptation of GRPO to multi-turn tasks. However, GRPO is widely reported to be unstable in this setting. To address this, RAGEN \citep{wang2025ragen} proposes StarPO-s, which uses proportional trajectory filtering, while GiGPO \citep{feng2025group} refines advantage estimation by combining state-level advantages obtained by merging identical states with trajectory-level advantages. MT-GRPO \citep{zeng2025reinforcing} further demonstrates the benefit of turn-level credit assignment. Nonetheless, these approaches are not fully general and often require manual tuning of turn-level credit weights. In this work, we reintroduce a learnable critic, as in PPO, but applied at the turn level, providing effective turn-level advantage estimates and mitigating the limitations of GRPO in multi-turn training.\footnote{Concurrent with our work, GEM \citep{liu2025gem} and ST-PPO \citep{li2025st} also identify limitations of GRPO in multi-turn settings and observed that turn-level PPO typically yields better performance, particularly on complex, long-horizon tasks.} 
\section{Framework}
In this section, we introduce our RL framework for multi-turn agentic LLM tasks.
Section~\ref{sec: llm mdp} presents the general LLM-as-MDP formulation, followed by Section~\ref{sec: token mdp}, which illustrates the token-MDP formulation and analyzes its limitations.
To address these issues, Section~\ref{sec: turn mdp} introduces the turn-MDP formulation.
For each formulation, we provide the corresponding GRPO and PPO algorithms under a unified framework.
Finally, Section~\ref{sec: rl algo compare} compares these algorithms in detail.

\begin{figure*}
    \centering
    \includegraphics[width=1.0\linewidth]{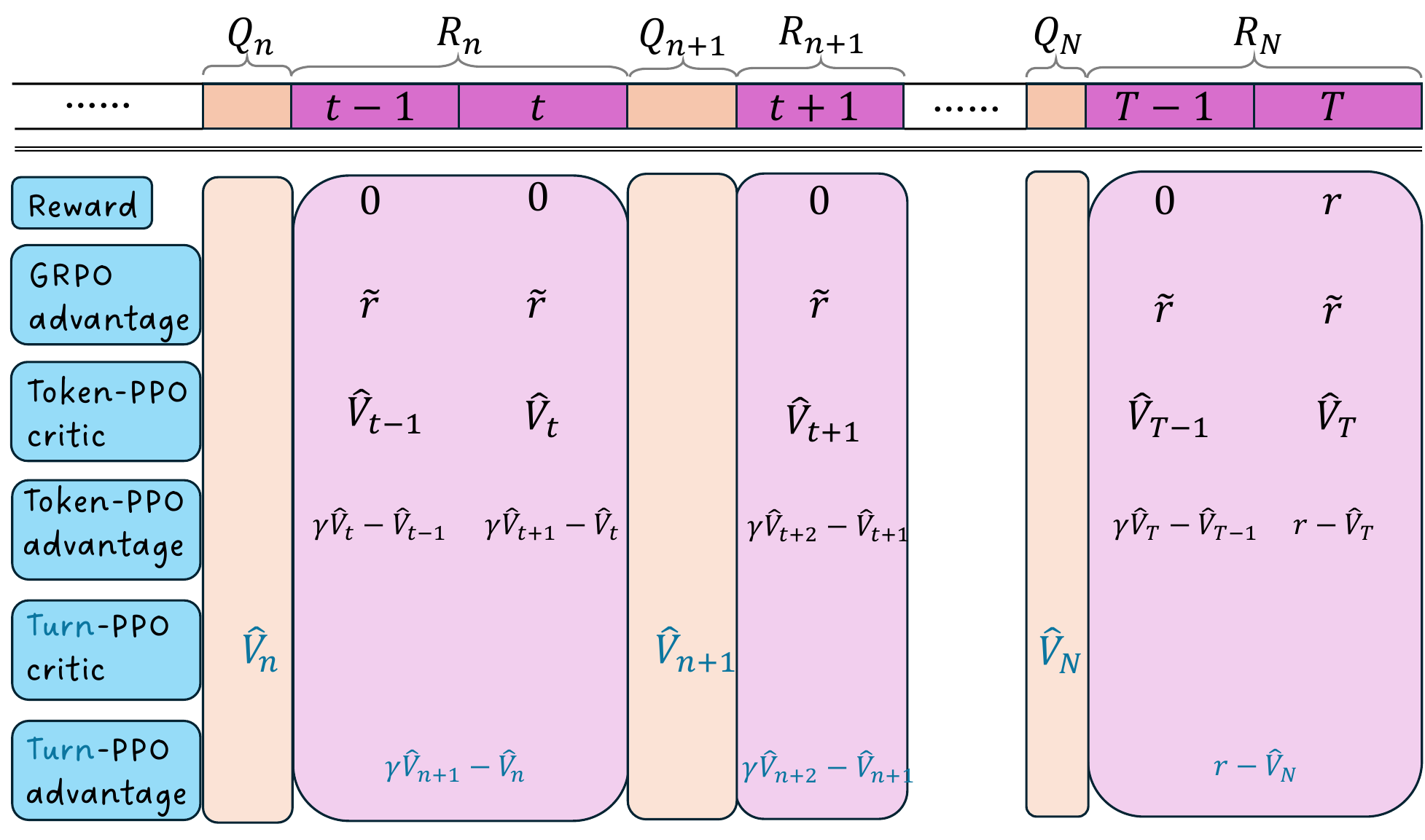}
    \caption{Comparison of advantage computation in GRPO, token-PPO, and turn-PPO. In turn-PPO, the state is defined as $s_n := \left(\oplus_{n'<n}(Q_{n'}, R_{n'})\right) \oplus Q_n$ and the action as $a_n := R_n$. For the critic in token-PPO and turn-PPO, the position of $\hat{V}_h$ in the figure indicates that it is conditioned on all tokens up to that point.}
    \label{fig:algorithm visulization}
    % \vspace{-4mm}
\end{figure*}

\subsection{Multi-turn LLM and MDP}
\label{sec: llm mdp}
\paragraph{Multi-turn LLMs}
We first clarify the multi-turn setting in LLM. For a full episode with $N\geq 1$  turns, each turn $n$ includes the query tokens from the environment denote as $Q_n$ and the LLM response tokens generated by LLM denoted as $R_n$. The $Q_n$ may include system prompts, user query, environment/tool output as the input to LLMs, and the $R_n$ is the output of LLMs. When $N=1$, we recover the single-turn setting. We illustrate in the following that the differences of the RL algorithms are how we formulate the MDP for the turns formed by $(Q_n, R_n)$.

\paragraph{LLM-as-MDP}
We model LLM training in RL as a Markov Decision Process (MDP) defined by $(\mathcal{S}, \mathcal{A}, H, \mathbb{P}, r)$, where $\mathcal{S}$ denotes the state space, $\mathcal{A}$ the action space, $H \in \mathbb{N}$ the horizon length, $\mathbb{P}$ the state transition probability measure, and $r: \mathcal{S} \times \mathcal{A} \rightarrow \mathbb{R}$ the reward function. Denote $\mathcal{D}$ to be the query set, $\pi_\theta$ the parameterized policy. In policy-based RL for LLMs, the goal is to learn a policy $\pi: \mathcal{S} \times \mathcal{A} \rightarrow [0,1]$ that maximizes the expected return:
% \begin{equation}
% \max_{\theta} \ \mathbb{E}_{s_1 \sim \mathcal{D}} \sum_{h=1}^H \mathbb{E}_{a_h \sim \pi_{\theta}(\cdot|s_h),\ s_{h+1} \sim \mathbb{P}(\cdot|s_h,a_h)} r(s_h, a_h).
% \end{equation}
\begin{equation}
\max_{\theta} \ \mathbb{E}_{s_1 \sim \mathcal{D}} 
\sum_{h=1}^H 
\mathbb{E}_{\substack{
a_h \sim \pi_{\theta}(\cdot|s_h),\\
s_{h+1} \sim \mathbb{P}(\cdot|s_h, a_h)
}} r(s_h, a_h).
\end{equation}

We solve this optimization via policy gradient methods such as REINFORCE \citep{williams1992simple}, PPO \citep{schulman2017proximal}, and GRPO \citep{shao2024deepseekmath}. In LLM training, both states and actions are represented by text tokens, and rewards are typically given only at sequence end, with optional KL shaping applied to intermediate tokens. In the following, we denote $\text{clip}_{\varepsilon}(x):= \text{clip}(x, 1-\varepsilon, 1+\varepsilon)$.

\subsection{Token-MDP}
\label{sec: token mdp}
In current widely-used token-MDP formulation, each time the action $a_h$ is just one token generated by LLM. We illustrate the objectives for single-turn setting and multi-turn setting respectively.

\paragraph{Single-turn settings}
In the single-turn setting, the model receives a single query and generates all answer tokens in one round, without intermediate tokens from external sources (e.g., tool outputs or follow-up user messages). 
% This can be viewed either as a bandit problem, where the entire response is one action ($H=1$), or as a standard MDP where each token is treated as an action ($H\geq 1$).  
Formally, at step $h$, the state $s_h$ consists of the query and the first $h-1$ generated tokens; the action $a_h$ is the $h$-th token; and the next state $s_{h+1}$ is obtained by concatenating $s_h$ with $a_h$, resulting in a deterministic transition $\mathbb{P}$. The current design of both PPO and GRPO view the response as actions of tokens. In this case, we here show the unified loss functions for PPO and GRPO in Table \ref{tab:objs}'s first line, with different choices of specific parts leading to each variant. We omit the KL part as that's not the focus of this work and can be added empirically. 
% Specifically, we solve the above reward maximization problem offline with importance sampling and clipping in PPO and GRPO with loss function shown in the first line of Table \ref{tab:objs}.

% \begin{align}
%     \max_{\theta} \ \mathbb{E}_{s_1 \sim \mathcal{D}, \ \{a^i\}_{i=1}^G \sim \pi_{\theta_t}(\cdot|s)} 
%     \frac{1}{G} \sum_{i=1}^G \frac{1}{|a^i|} \sum_{h=1}^{|a^i|} 
%     \min\!\left(
%         r^i_{h}(\theta) \, \hat{A}^i_h, \ 
%         \text{clip}_{\varepsilon}\left(r^i_{h}(\theta)\right) \, \hat{A}^i_h
%     \right),
% \end{align}

% where $r^i_{h}(\theta) = \frac{\pi_{\theta}(a^i_h|s^i_h)}{\pi_{\theta_t}(a^i_h|s^i_h)}$ is the importance ratio of the $h$-th token. 
For PPO, we do not need multiple sampling for one initial query so $G$ is usually set to be 1, while for GRPO, $G>1$ since we need to compute group relative advantage. For PPO, $\hat{A}_h^i$ is the token-level advantage estimated via Generalized Advantage Estimation (GAE) using a learned critic model. For GRPO, the advantage is constant across tokens, i.e., $\hat{A}_h^i := \hat{A}^i$, where $\hat{A}^i$ is computed by normalizing the final sequence rewards over the $G$ rollouts for the same query.

\paragraph{Multi-turn setting} 
In the multi-turn setting, external tokens are injected into the context during the interaction, such as tool outputs or follow-up user queries.  
A straightforward extension from the single-turn case is to retain the same implementation but mask out the loss for environment tokens, keeping only the loss on LLM-generated tokens \citep{wei2025webagent,wang2025ragen}. Denote $a_n^i$ to be the LLM response in $n$-th turn at the $i$-th sample of the query $s_1$, and the similar representation goes for probability ratio $r^i_{n,h}$ and advantage $\hat{A}^i_{n,h}$. The unified objective for GRPO and PPO is shown in Table \ref{tab:objs}'s second line.

% \begin{align}
%     \max_{\theta} \ \mathbb{E}_{s_1 \sim \mathcal{D}, \ \{a^i\}_{i=1}^G \sim \pi_{\theta_t}(\cdot|s)} 
%     \frac{1}{G} \sum_{i=1}^G \frac{1}{\sum_{n=1}^{N_i}|a_n^i|} \sum_{n=1}^{N_i}\sum_{h=1}^{|a_n^i|} 
%     \min\!\left(
%         r^i_{n,h}(\theta) \, \hat{A}^i_{n,h}, \ 
%         \text{clip}_{\varepsilon}\left(r^i_{n,h}(\theta)\right) \, \hat{A}^i_{n,h}
%     \right).
% \end{align}

Though empirically easily to implement, we point out that there may be fundamental problems of this objective.
The MDP formulation for this multi-turn setting diverges from the single-turn case, leading to a ``\textbf{state representation misalignment}.'' 
% lihongli: the terminology seems strange.
This introduces additional noise into the training of the critic $V(\cdot)$, making it harder to estimate accurate advantages. In a token-level MDP, state transitions are not continuous in the multi-turn setting: within a response, the next state is simply the current state with one token appended, whereas across responses, the next state incorporates an entire block of tokens from the environment output. These heterogeneous transition patterns cause the critic to regress toward an averaged value, reducing the fidelity of state values and ultimately degrading the quality of the computed advantages.

\subsection{Turn-MDP}
\label{sec: turn mdp}
Rather than taking a direct empirical extension at the implementation level, we instead adopt a different MDP formulation. In fact, the multi-turn interaction setting naturally induces an MDP, since environment interactions and state transitions unfold across turns. Concretely, we define the state $s_h$ as the full interaction history up to the $(h-1)$-th round together with the current $h$-th query, and the action $a_h$ as the LLM’s response at round $h$. Thus, unlike token-level actions, here each $a_h$ corresponds to the entire response in a given turn. This formulation yields a more uniform representation: every $s_h$ is the complete history concatenated with the current query, and every $a_h$ is the full response to that query. We specify the objectives for single-turn setting and multi-turn setting in Table \ref{tab:objs}'s third and fourth lines respectively. 

\paragraph{Single-turn setting}
% \begin{align}
%     \max_{\theta} \ \mathbb{E}_{s_1 \sim \mathcal{D}, \ \{a^i\}_{i=1}^G \sim \pi_{\theta_t}(\cdot|s)} 
%     \frac{1}{G} \sum_{i=1}^G \frac{1}{|a^i|} 
%     \min\!\left(
%         r^i(\theta) \, \hat{A}^i, \ 
%         \text{clip}_{\varepsilon}\left(r^i(\theta)\right) \, \hat{A}^i
%     \right),
% \end{align}
% where $r^i(\theta)=\frac{\pi_{\theta}(a^i|s^i)}{\pi_{\theta_t}(a^i|s^i)}=\prod_{h=1}^{|a^i|}\frac{\pi_\theta(a_h^i|s_h^i)}{\pi_{\theta_t}(a_h^i|s_h^i)}$ is the importance ratio of the whole response. 
Empirically, we can also take the geometric average of the ratio to improve stability, \emph{i.e.}, using ${r^i}^{1/|a^i|}$ instead. We note here that this version of GRPO with turn-MDP and geometric average recovers GSPO \citep{zheng2025group} as the special case.
\paragraph{Multi-turn setting}
% \begin{align}
%     \max_{\theta} \ \mathbb{E}_{s_1 \sim \mathcal{D}, \ \{a^i\}_{i=1}^G \sim \pi_{\theta_t}(\cdot|s)} 
%     \frac{1}{G} \sum_{i=1}^G \frac{1}{\sum_{n=1}^N|a_n^i|} \sum_{n=1}^N 
%     \min\!\left(
%         r^i_{n}(\theta) \, \hat{A}^i_{n}, \ 
%         \text{clip}_{\varepsilon}\left(r^i_{n}(\theta)\right) \, \hat{A}^i_{n}
%     \right),
% \end{align}
% where $r_n^i(\theta)=\frac{\pi_{\theta}(a_n^i|s_n^i)}{\pi_{\theta_t}(a_n^i|s_n^i)}=\prod_{h=1}^{|a_n^i|}\frac{\pi_\theta(a_{n,h}^i|s_{n,h}^i)}{\pi_{\theta_t}(a_{n,h}^i|s_{n,h}^i)}$ is the importance ratio of response in the $n$-th turn. 
For GRPO-style advantage estimation, $\hat{A}_{n}^i:=\hat{A}^i$ can be empirically computed as the normalized final rewards irrelevant to turn and token, similar as the multi-turn token-MDP setting. For PPO, $\hat{A}_{n}^i$ is well-defined by turn-level GAE advantage. The difference is that the learned critic model is in the turn level. We name this version of PPO ``\textbf{turn-PPO}''. 
Besides the actor loss defined above, the critic is optimized using a turn-level PPO-style value objective. 
Specifically, we initialize the critic from a pre-trained LLM but attach a separate value head. 
The value loss is given by:
\begin{align}
    \min_{\phi}~
    \mathbb{E}_{\substack{
        s_1 \sim \mathcal{D},\\
        \{a^i\}_{i=1}^G \sim \pi_{\theta_t}(\cdot|s_i),\\
        s_{n+1}^i \sim \mathcal{P}(\cdot|s_n^i,a_n^i)
    }}
    \frac{1}{G}\sum_{i=1}^G \frac{1}{N_i}\sum_{n=1}^{N_i}
    \frac{1}{2}\big(V_\phi(s_n^i) - \hat{R}_{n}^i\big)^2,
\end{align}
where $\hat{R}_{n}^i$ denotes the cumulative discounted return from turn $n$ onward.

\begin{table*}[t]
\centering
\small
\renewcommand{\arraystretch}{1.4}
\begin{tabular}{p{1.0\linewidth}}
\toprule
\textbf{Token-Single:} $\displaystyle
\max_{\theta} \mathbb{E}_{\substack{s_1 \sim \mathcal{D}\\\{a^i\}\!\sim\!\pi_{\theta_t}}}
\frac{1}{G}\!\sum_{i=1}^G \frac{1}{|a^i|}\!\sum_{h=1}^{|a^i|}
\mathrm{MC}_{\varepsilon}\!\big(r^i_{h}(\theta), \hat{A}^i_h\big),
\ \text{where } r^i_{h}(\theta)=\frac{\pi_{\theta}(a^i_h|s^i_h)}{\pi_{\theta_t}(a^i_h|s^i_h)}.$ \\[0.4em]
\midrule

\textbf{Token-Multi:} $\displaystyle
\max_{\theta} \mathbb{E}_{\substack{s_1 \sim \mathcal{D}\\\{a^i\}\!\sim\!\pi_{\theta_t}\\ s_{n+1}^i \sim \mathcal{P}}}
\frac{1}{G}\!\sum_{i=1}^G \frac{1}{|a^i|}\!\sum_{n=1}^{N_i}\!\sum_{h=1}^{|a_n^i|}
\mathrm{MC}_{\varepsilon}\!\big(r^i_{n,h}(\theta), \hat{A}^i_{n,h}\big),
\ \text{where } r^i_{n,h}(\theta)=\frac{\pi_{\theta}(a^i_{n,h}|s^i_{n,h})}{\pi_{\theta_t}(a^i_{n,h}|s^i_{n,h})}.$ \\[0.4em]
\midrule

\textbf{Turn-Single:} $\displaystyle
\max_{\theta} \mathbb{E}_{\substack{s_1 \sim \mathcal{D}\\\{a^i\}\!\sim\!\pi_{\theta_t}}}
\frac{1}{G}\!\sum_{i=1}^G \frac{1}{|a^i|}
\mathrm{MC}_{\varepsilon}\!\big(r^i(\theta), \hat{A}^i\big),
\ \text{where } r^i(\theta)=\frac{\pi_{\theta}(a^i|s^i)}{\pi_{\theta_t}(a^i|s^i)}=\prod_{h=1}^{|a^i|}\frac{\pi_{\theta}(a^i_h|s^i_h)}{\pi_{\theta_t}(a^i_h|s^i_h)}.$ \\[0.4em]
\midrule

\textbf{Turn-Multi:} $\displaystyle
\max_{\theta} \mathbb{E}_{\substack{s_1 \sim \mathcal{D}\\\{a^i\}\!\sim\!\pi_{\theta_t}\\ s_{n+1}^i \sim \mathcal{P}}}
\frac{1}{G}\!\sum_{i=1}^G \frac{1}{|a^i|}\!\sum_{n=1}^{N_i}
\mathrm{MC}_{\varepsilon}\!\big(r^i_{n}(\theta), \hat{A}^i_{n}\big),
\ \text{where } r^i_{n}(\theta)=\frac{\pi_{\theta}(a^i_{n}|s^i_{n})}{\pi_{\theta_t}(a^i_{n}|s^i_{n})}=\prod_{h=1}^{|a^i_n|}\frac{\pi_{\theta}(a^i_{n,h}|s^i_{n,h})}{\pi_{\theta_t}(a^i_{n,h}|s^i_{n,h})}.$ \\[0.2em]
\bottomrule
\end{tabular}
\caption{Comparison of PPO objectives for Token-/Turn-level MDPs in Single- and Multi-turn settings. Here, $\mathrm{MC}_{\varepsilon}(r,A):=\min\!\big(r\,A,\ \text{clip}_{\varepsilon}(r)\,A\big)$ denotes the PPO min-with-clipping operator.}
\label{tab:objs}
\vspace{-2mm}
\end{table*}

\subsection{RL algorithm comparison}
\label{sec: rl algo compare}

For better understanding the differences, we visualize the differences in advantage computation between GRPO, 
token-PPO, and turn-PPO in Figure \ref{fig:algorithm visulization}. 
Suppose the trajectory contains $N$ turns and $T$ tokens in total. 
For clarity, we only consider the final reward $r$ at the end of the trajectory 
and omit token-level shaping rewards (e.g., KL penalties).

For \textbf{GRPO}, the advantage $\tilde r$ is obtained by normalizing rewards across 
multiple trajectories for the same query and then assigning this normalized value 
to every token. Each token is treated as an individual action, so clipping 
and loss aggregation are performed at the token level.

For \textbf{token-PPO} and \textbf{turn-PPO}, we use generalized advantage estimation (GAE) \citep{murphy2024reinforcement} defined as:
\begin{align*}
\delta_h &= r_h + \gamma v_{h+1} - v_h, \\
A_h &= \delta_h + \gamma \lambda \delta_{h+1} + \cdots + (\gamma \lambda)^{H-(h+1)} \delta_{H-1} \\
&= \delta_h + \gamma \lambda A_{h+1}.
\end{align*}
Here, $\gamma$ is the reward discount factor across timesteps in the MDP, while $\lambda \in [0,1]$ controls the bias-variance trade-off: larger values reduce bias but increase variance. For simplicity, we only show the advantage for a single step, \emph{i.e.}, $\delta_h$, in the Figure.
\textbf{Token-PPO} uses the critic's value prediction at each token as $V_t$ and updates according to the Bellman equation. The advantage at token $t$ is computed as $\gamma V_{t+1} - V_t$ for intermediate tokens, and $r - V_T$ for the final token.
\textbf{Turn-PPO}, in contrast, operates at the \emph{turn level}: both actor and critic losses are defined per turn. The value function is given by the critic's output at the \emph{last token} of each turn (query or environment output). The entire LLM response for that turn is considered the action, so the computed advantage applies to the whole response. Formally, the advantage is $\gamma V_{n+1} - V_n$ for intermediate turns and $r - V_N$ for the final turn. Clipping is also performed at the response (turn) level.
We note that turn-level PPO incurs the same computational cost as token-level PPO, as the only difference lies in computing the loss at the turn level, while the forward and backward passes remain identical.

\section{Experiments}
Section~\ref{sec: setup} introduces the experimental setup.
In Section~\ref{sec:grpo exp}, we analyze GRPO and its variants, demonstrating their failure modes and highlighting the challenges of applying GRPO directly in multi-turn settings.
Section~\ref{sec:ppo exp} then focuses on PPO-based algorithms, showing the advantages of turn-PPO and providing improved training guidelines for PPO in LLMs.
\subsection{Setup}
\label{sec: setup}
\paragraph{Data}
Our experiments focus on multi-turn environments, specifically WebShop \citep{yao2022webshop} and Sokoban \citep{junghanns2001sokoban}, both of which require sequential decision-making and long-horizon reasoning. In WebShop, the model receives a user query and must complete a sequence of actions: searching for products, selecting relevant items, refining attributes like color or size, and finalizing the purchase. The process frequently involves iterative searches and navigating across multiple pages, testing the model’s ability to maintain coherent goal-directed behavior over many steps. In Sokoban, the model faces a spatial planning task, pushing boxes to target locations in a grid world, where each move has irreversible consequences, incurs a time penalty, and only a sparse terminal reward is provided. These environments capture different challenges of multi-step alignment and planning. Further dataset examples are provided in Appendix \ref{sec: example}.

\begin{figure*}[h!]
    \centering
    \includegraphics[width=1.0\linewidth]{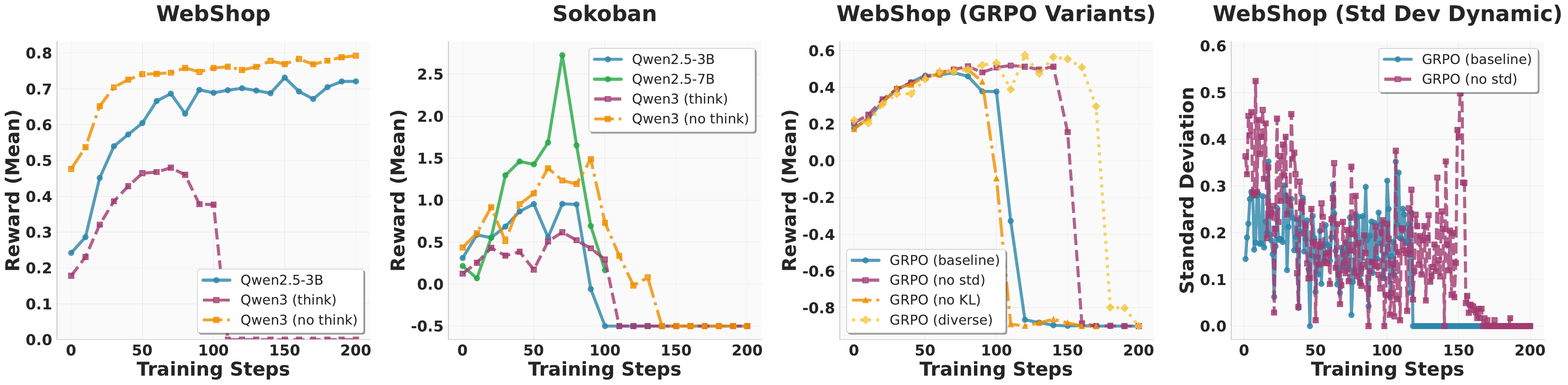}
    \caption{In the first two plots, we show GRPO validation reward curves during training on Webshop and Sokoban for Qwen2.5 and Qwen3. In the third one, we show rewards for GRPO and its variants with respect to std dev, KL, and batch size diversity. In the last one, we show evolution of standard deviation throughout training.}
    \label{fig:grpo main}
    % \vspace{-5mm}
\end{figure*}

\paragraph{Algorithm}
For the token-MDP formulation, we adopt GRPO and token-PPO as baselines. For our turn-MDP formulation, we focus on turn-PPO as the main algorithm, since it offers clearer design insights. In Section \ref{sec:grpo exp}, we examine the common approach of extending GRPO to multi-turn settings under token-MDP, highlighting issues such as instability. In Section \ref{sec:ppo exp}, we present a systematic comparison of GRPO, token-PPO, and turn-PPO across tasks and models with varying reasoning behaviors, demonstrating the advantages of our turn-PPO method. Our training codes are based on RAGEN \citep{wang2025ragen}, with turn-PPO implemented.

\paragraph{Model}
Our RL experiments are conducted using Qwen2.5-3B and Qwen2.5-7B \citep{team2024qwen2}, as well as Qwen3-1.7B \citep{yang2025qwen3} as the base models. For Qwen2.5, reasoning is always enabled. For Qwen3, we evaluate both reasoning-enabled and reasoning-disabled settings. This choice is motivated by the observation that Qwen3’s reasoning tends to be excessively long, which creates two issues: (1) overthinking for our tasks, and (2) extended reasoning chains combined with long RL trajectories make training unstable, offering little improvement and often leading to collapse.

\subsection{GRPO failure investigation}
\label{sec:grpo exp}
We apply GRPO with a token-MDP formulation for both tasks, following the default implementation in RAGEN and other frameworks. For each query in the dataset, we sample multiple trajectories and compute advantage estimates using normalized rewards. However, this setup exhibits significant instability: in both tasks, training often collapses abruptly, particularly in the long-reasoning setting with Qwen3. The training curves in Figure \ref{fig:grpo main} illustrate these crashes across both tasks for both Qwen2.5 and Qwen3.

To gain deeper insight into GRPO training behavior, we investigate the causes of its failures and explore potential improvements. Our analysis shows that these failures are systematic and cannot be resolved by variants within the GRPO framework. In particular, we focus on Qwen3 with reasoning on the WebShop dataset, where training crashes occur most frequently.

Our first hypothesis attributes the instability to potential entropy collapse \citep{wang2025ragen}, where GRPO’s normalized rewards cause large fluctuations in the advantage estimates. To test this, we \textbf{remove the standard deviation term} in reward normalization from vanilla GRPO, a modification that has been reported to help in methods such as Reinforce-Rej \citep{xiong2025minimalist} and DR.GRPO \citep{liu2025understanding}. However, our results show no mitigation. Moreover, visualizing the average reward standard deviation during RL training further confirms that entropy collapse may not be the primary issue as shown in Figure \ref{fig:grpo main}.

Second, we experiment with \textbf{removing the KL term} in GRPO, as the model distribution tends to diverge significantly in long-reasoning settings such as Qwen3, and enforcing KL regularization in such cases may overly constrain learning. This choice is also supported by findings in DAPO \citep{yu2025dapo}. Nevertheless, this modification also shows only minimal effect and does not prevent the crash, as illustrated in Figure \ref{fig:grpo main}.

Next, we \textbf{enhance the diversity} of each training batch to mitigate overfitting on limited samples. Specifically, while keeping the total number of rollout samples each time fixed, we decrease the number of rollouts per question and increase the number of distinct questions. This adjustment provides a slight improvement, but merely delays the crash and fails to fundamentally resolve the issue or improve overall performance.

Consequently, all of the above modifications to the original GRPO framework still fail to
prevent training collapse. This highlights a fundamental limitation of directly applying GRPO
to the multi-turn setting under a token-MDP formulation. We attribute this failure to two key
factors:
\begin{enumerate}
    \item The use of a uniform advantage across all turns, which overlooks the varying levels of difficulty and distinct characteristics that cause different turns to contribute unequally.
    \item The high variance of sample-based advantage estimates, which is amplified in multi-turn settings with dynamic or partially observed environments.
\end{enumerate}
Together, these factors promote over-training on easier turns and eventually cause model collapse.
This issue is especially pronounced in multi-turn long-reasoning scenarios, where turn boundaries are well defined and exhibit substantially greater heterogeneity than in single-turn settings. Moreover, long reasoning trajectories contain a higher proportion of low-entropy tokens, which makes the model more prone to collapse when trained on them \citep{wang2025ragen}.

\subsection{Token-PPO and turn-PPO}
\label{sec:ppo exp}

\begin{figure*}[h!]
    \centering
    \includegraphics[width=1.0\linewidth]{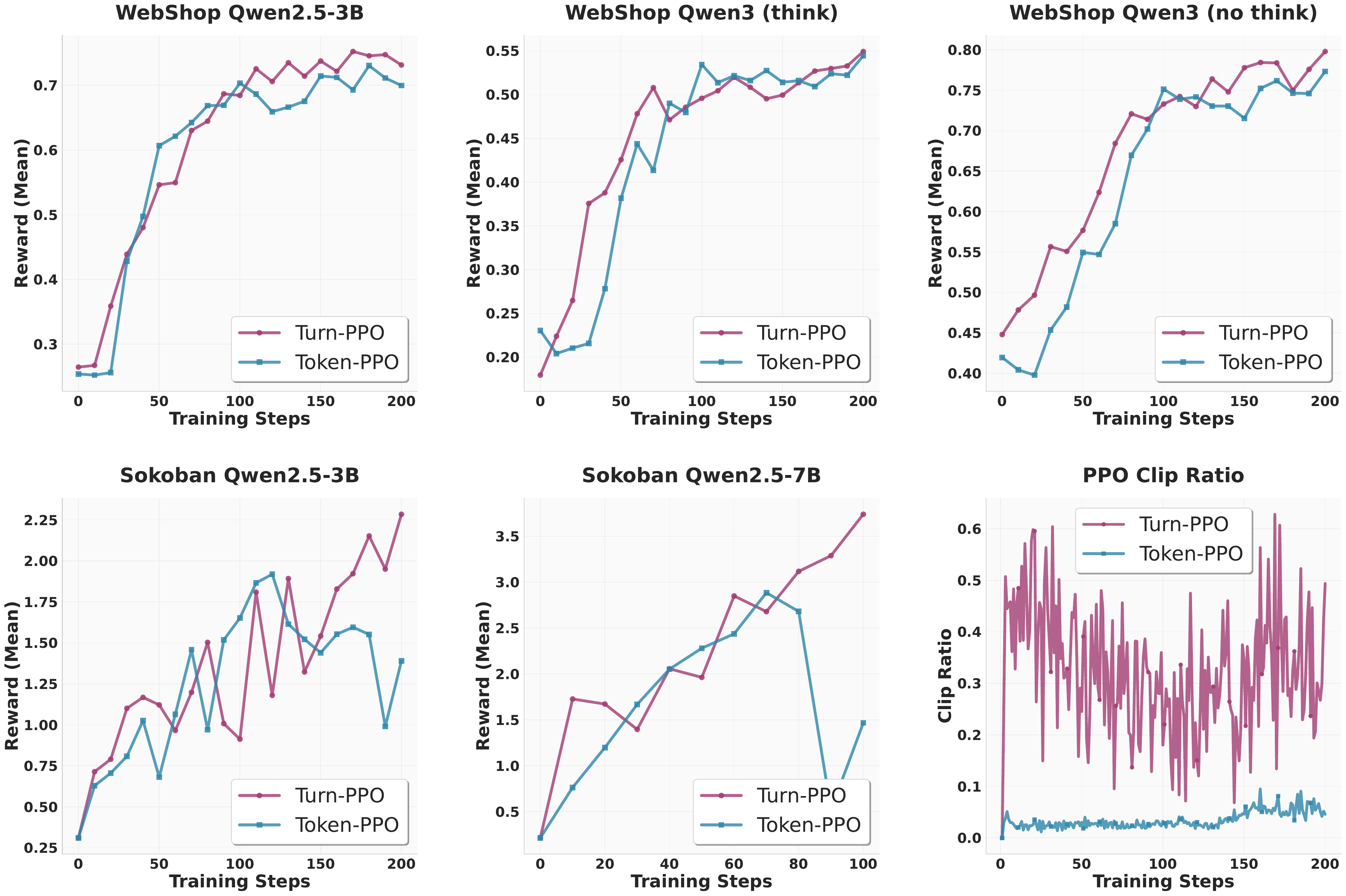}
    \caption{Comparison of turn-PPO and token-PPO in mean reward across multiple settings and and clipping ratio for Sokoban. Turn-PPO shows superior performance in most settings, highlighting the benefit of turn-level advantage estimation.}
    \label{fig:ppo results}
    \vspace{-2mm}
\end{figure*}

We then apply PPO-based algorithms to these tasks, using two variants: token-PPO, the baseline implementation based on a token-level MDP, and turn-PPO, our proposed variant formulated on a turn-level MDP. Experimental results show that both PPO-based methods substantially outperform GRPO on multi-turn tasks. Furthermore, turn-PPO demonstrates improved training stability and achieves superior performance in most cases. As illustrated in Figure \ref{fig:ppo results}, PPO effectively mitigates the training collapse observed with GRPO, providing a more stable and robust learning-based advantage estimation strategy.

\paragraph{Clipping} We visualize the token clipping ratio for token-PPO and turn-PPO for Sokoban tasks in Figure~\ref{fig:ppo results}. Turn-PPO shows a much higher clipping ratio because clipping is applied to the entire response at the turn level: if the policy distribution changes too drastically, the whole turn is clipped. This prevents the model from updating on turns with large policy shifts, avoiding unstable gradient steps and enabling smoother, more reliable training.

\paragraph{Model backbone} 
For all tasks, we initially experiment with both Qwen2.5 and Qwen3, enabling thinking by default. We observe that Qwen2.5 produces reasoning of appropriate length, which improves over the course of training. In contrast, Qwen3 generates excessively long and often unnecessary reasoning at the beginning, and this behavior does not improve during training, resulting in no gain in final reward. Consequently, we also evaluate Qwen3 with thinking disabled, which performs significantly better—surpassing Qwen2.5. We hypothesize that the mismatch between Qwen3’s thinking distribution and our task setup introduces instability, causing GRPO training to collapse. PPO generally mitigates this issue, and our proposed turn-PPO further enhances training stability and efficiency. We show trajectory examples for WebShop in Appendix \ref{sec: example}.

\paragraph{GRPO \emph{v.s.} PPO}
We present the comparison between GRPO and PPO variants in Table~\ref{tab:grpo ppo}. As shown, Token-PPO performs comparably to GRPO or stabilizes its previously unstable training, while our Turn-PPO achieves further improvements.

\begin{table}[h!]
\centering
\caption{Average reward comparison across GRPO, Token-PPO, and Turn-PPO on WebShop and Sokoban benchmarks. “Crash” indicates failed RL training runs.}
\label{tab:grpo ppo}
\resizebox{1.0\linewidth}{!}{%
\begin{tabular}{llccc}
\toprule
\textbf{Environment} & \textbf{Model} & \textbf{GRPO} & \textbf{Token-PPO} & \textbf{Turn-PPO} \\
\midrule
\multirow{3}{*}{WebShop} 
 & Qwen2.5-3B & 0.72 & 0.73 & \textbf{0.75} \\
 & Qwen3-1.7B (no think) & 0.78 & 0.77 & \textbf{0.80} \\
 & Qwen3-1.7B (think) & Crash & 0.54 & \textbf{0.55} \\
\midrule
\multirow{2}{*}{Sokoban} 
 & Qwen2.5-3B & Crash & 1.93 & \textbf{2.29} \\
 & Qwen2.5-7B & Crash & 2.90 & \textbf{3.74} \\
\bottomrule
\end{tabular}%
}
\end{table}

\subsection{Ablation studies: PPO recipe}
We conduct ablation studies on the hyperparameters of the PPO-based algorithms. Given the large number of hyperparameters and their high sensitivity to final performance, our goal is to identify the optimal training strategy for PPO-based methods. Our ablation focuses on the more challenging setting with Qwen3 with long reasoning.

\paragraph{Learning rate} 
PPO employs two separate learning rates for the actor and the critic respectively. We find both to be highly sensitive: even slight adjustments can cause training instability or complete failure. To provide meaningful learning signals, the critic’s learning rate must be approximately 5--10$\times$ higher than the actor’s; otherwise, training can stagnate or diverge. Given this sensitivity, we set the actor and critic learning rates to $1 \times 10^{-6}$ and $1 \times 10^{-5}$, respectively.

\paragraph{Batch size, epoch, and batch diversity} 
In PPO, the effective batch size is determined by two components: the number of rollout samples $B_R$ collected per iteration and the minibatch size $B_M$ used during updates. Additionally, one must select the number of epochs $E$ to iterate over each rollout dataset. A related hyperparameter is the group size $G$, defined as the number of rollouts for each distinct sample. Since rollout generation is the primary bottleneck in LLM RL training, we keep $B_R$ fixed and tune $G$, $B_M$, and $E$ to balance sample efficiency and training stability.

These hyperparameters are shared between GRPO and PPO, so we study them jointly. 
For GRPO, we begin with the baseline configuration $(B_R, G, B_M, E)=(32, 16, 32, 1)$, where each rollout collects $32$ trajectories consisting of $2$ distinct questions with $16$ trajectories each. 
The mini-batch size is set to $32$, and we use a single epoch ($E=1$), meaning the collected data is used exactly once for model updates. 
For PPO, the baseline is $(B_R, G, B_M, E)=(32, 1, 32, 1)$, where the $32$ trajectories each correspond to a different question sampled once. 
This highlights a key advantage of PPO over GRPO: for the same total number of rollouts, PPO sees a wider variety of problems because it does not require multiple trajectories per question.

\begin{figure}[h!]
    \centering
    \includegraphics[width=1.0\linewidth]{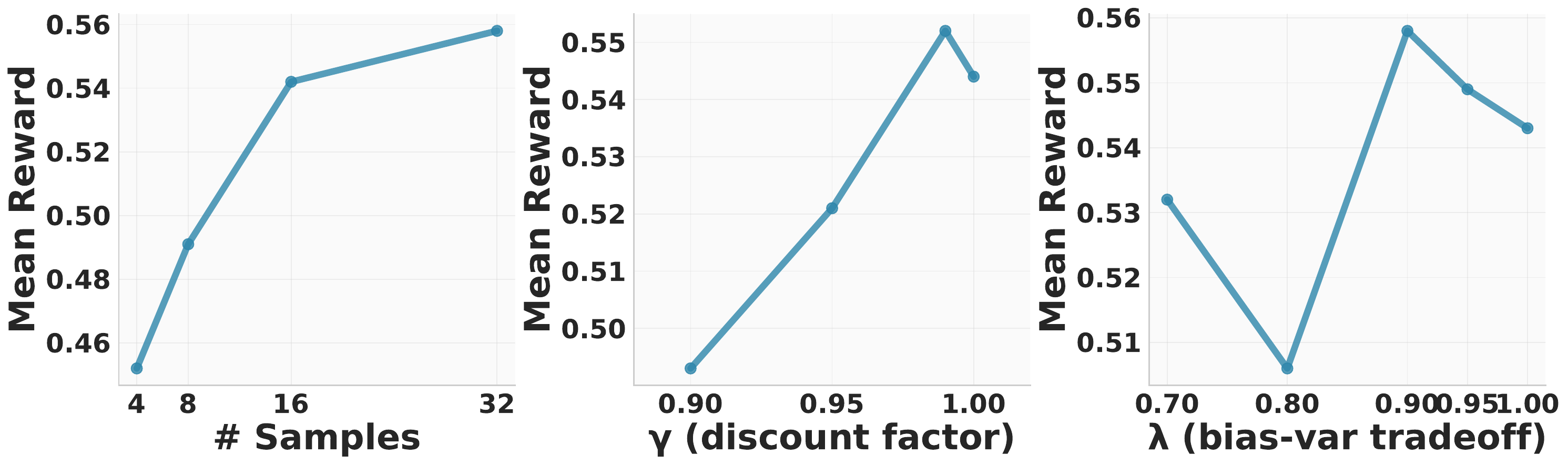}
    \caption{Ablation studies on (left) number of diverse samples in a batch, (middle) discount factor $\gamma$, and (right) bias–variance trade-off parameter $\lambda$, showing their impact on mean reward using WebShop and Qwen3 with reasoning.}
    \label{fig:ablate div}
    % \vspace{-2mm}
\end{figure}

\begin{itemize}
    \item We first ablate the \textbf{batch diversity} $G$ for both GRPO and PPO and observe distinct behaviors. 
    For GRPO, when training remains stable, increasing the number of samples per question improves performance by yielding more accurate advantage estimates. 
    For PPO, in contrast, using a single sample per question gives the best results. 
    We attribute this to the fact that higher problem diversity helps the critic generalize better and prevents overfitting to individual questions. 
    The trends with respect to $G$ for PPO are illustrated in Figure~\ref{fig:ablate div}.
    
    \item We then ablate $(B_M, E)$, which correspond to the classical \textbf{batch size and number of epochs} in standard deep learning. 
    In the baseline setting, both GRPO and PPO perform only one model update per rollout, representing a strict online RL regime. 
    Because rollouts are the most expensive part of LLM RL training, we explore more sample-efficient offline strategies that reuse collected data multiple times. 
    Specifically, we experiment with smaller minibatch sizes $B_M$ and increased epochs $E$. 
    Our results suggest it is preferable to first decrease $B_M$ rather than increase $E$, since excessively reusing the same data across many epochs risks overfitting. 
\end{itemize}
% \vspace{-3.5mm}

\paragraph{Generalized advantage estimation: decay $\gamma$ and bias-variance trade-off $\lambda$} 
We next discuss the two key hyperparameters in generalized advantage estimation (GAE) \citep{murphy2024reinforcement} used in PPO. 

Both parameters are critical for stable and effective advantage estimation in PPO.
We emphasize that our turn-PPO allows more flexible tuning of $\gamma$ and $\lambda$, yielding better performance. 
However, in token-level PPO both must be fixed at $1.0$. 
The reason is that in turn-MDP, these parameters operate at the turn level, so a change such as from $1.0$ to $0.95$ only slightly affects the first turn. 
In contrast, in token-MDP where trajectories may contain thousands of tokens, even a tiny reduction makes early tokens effectively invisible, causing training to diverge. 
Therefore, we report ablations only for turn-PPO in Figure~\ref{fig:ablate div}, where $\gamma=0.99$ and $\lambda=0.9$ are found to be a relatively stable and optimal choice.

The PPO recipe is summarized as follows:

\begin{findingbox}[To-Go List]
\begin{enumerate}[leftmargin=*, itemsep=4pt, topsep=2pt, label=\arabic*.]
    \item The learning rate is very sensitive, with the critic requiring a larger value than the actor.
    \item GRPO benefits from a larger number of rollouts per sample, whereas PPO performs better with greater sample diversity within each batch.
    \item Training with more minibatch updates per rollout is preferable to increasing the number of epochs.
    \item Turn-PPO allows flexible tuning of $\gamma$ and $\lambda$, consistently yielding improved results.
\end{enumerate}
\end{findingbox}

\section{Conclusion}
In this paper, we investigate RL algorithms for multi-turn agentic LLMs.
Within a unified framework, we systematically identify the limitations of existing GRPO and PPO algorithms in multi-turn settings, both theoretically and empirically, across multiple tasks and model scales.
To address these issues, we propose an improved PPO variant based on a turn-MDP formulation, which stabilizes training and better captures turn-level credit assignment.
Finally, we provide practical training guidelines for PPO-based algorithms, offering practical insights for future research and applications.

\newpage
\section*{Limitations}
Our study focuses on a detailed investigation of effective RL training algorithms and optimization strategies for multi-turn agents. As a result, we only evaluate our methods on two representative datasets, one for web-based agents and one for embodied agents. In the embodied setting, we primarily consider text-simulated environments. Future work should extend our approach to real-world web agents equipped with richer tool use and more complex decision-making, as well as to embodied scenarios involving physical interactions.

\bibliography{custom}

\begin{thebibliography}{39}
\providecommand{\natexlab}[1]{#1}

\bibitem[{Ahmadian et~al.(2024)Ahmadian, Cremer, Gall{\'e}, Fadaee, Kreutzer, Pietquin, {\"U}st{\"u}n, and Hooker}]{ahmadian2024back}
Arash Ahmadian, Chris Cremer, Matthias Gall{\'e}, Marzieh Fadaee, Julia Kreutzer, Olivier Pietquin, Ahmet {\"U}st{\"u}n, and Sara Hooker. 2024.
\newblock Back to basics: Revisiting reinforce style optimization for learning from human feedback in llms.
\newblock \emph{arXiv preprint arXiv:2402.14740}.

\bibitem[{Ethayarajh et~al.(2024)Ethayarajh, Xu, Muennighoff, Jurafsky, and Kiela}]{ethayarajh2024kto}
Kawin Ethayarajh, Winnie Xu, Niklas Muennighoff, Dan Jurafsky, and Douwe Kiela. 2024.
\newblock Kto: Model alignment as prospect theoretic optimization.
\newblock \emph{arXiv preprint arXiv:2402.01306}.

\bibitem[{Feng et~al.(2025)Feng, Xue, Liu, and An}]{feng2025group}
Lang Feng, Zhenghai Xue, Tingcong Liu, and Bo~An. 2025.
\newblock Group-in-group policy optimization for llm agent training.
\newblock \emph{arXiv preprint arXiv:2505.10978}.

\bibitem[{Guo et~al.(2025)Guo, Yang, Zhang, Song, Zhang, Xu, Zhu, Ma, Wang, Bi et~al.}]{guo2025deepseek}
Daya Guo, Dejian Yang, Haowei Zhang, Junxiao Song, Ruoyu Zhang, Runxin Xu, Qihao Zhu, Shirong Ma, Peiyi Wang, Xiao Bi, and 1 others. 2025.
\newblock Deepseek-r1: Incentivizing reasoning capability in llms via reinforcement learning.
\newblock \emph{arXiv preprint arXiv:2501.12948}.

\bibitem[{Gur et~al.(2023)Gur, Furuta, Huang, Safdari, Matsuo, Eck, and Faust}]{gur2023real}
Izzeddin Gur, Hiroki Furuta, Austin Huang, Mustafa Safdari, Yutaka Matsuo, Douglas Eck, and Aleksandra Faust. 2023.
\newblock A real-world webagent with planning, long context understanding, and program synthesis.
\newblock \emph{arXiv preprint arXiv:2307.12856}.

\bibitem[{Hu et~al.(2025)Hu, Liu, Xu, and Shen}]{hu2025reinforce++}
Jian Hu, Jason~Klein Liu, Haotian Xu, and Wei Shen. 2025.
\newblock Reinforce++: An efficient rlhf algorithm with robustness to both prompt and reward models.
\newblock \emph{arXiv preprint arXiv:2501.03262}.

\bibitem[{Huang et~al.(2022)Huang, Abbeel, Pathak, and Mordatch}]{huang2022language}
Wenlong Huang, Pieter Abbeel, Deepak Pathak, and Igor Mordatch. 2022.
\newblock Language models as zero-shot planners: Extracting actionable knowledge for embodied agents.
\newblock In \emph{International conference on machine learning}, pages 9118--9147. PMLR.

\bibitem[{Jin et~al.(2025)Jin, Zeng, Yue, Yoon, Arik, Wang, Zamani, and Han}]{jin2025search}
Bowen Jin, Hansi Zeng, Zhenrui Yue, Jinsung Yoon, Sercan Arik, Dong Wang, Hamed Zamani, and Jiawei Han. 2025.
\newblock Search-r1: Training llms to reason and leverage search engines with reinforcement learning.
\newblock \emph{arXiv preprint arXiv:2503.09516}.

\bibitem[{Junghanns and Schaeffer(2001)}]{junghanns2001sokoban}
Andreas Junghanns and Jonathan Schaeffer. 2001.
\newblock Sokoban: Enhancing general single-agent search methods using domain knowledge.
\newblock \emph{Artificial Intelligence}, 129(1-2):219--251.

\bibitem[{Kazemnejad et~al.(2024)Kazemnejad, Aghajohari, Portelance, Sordoni, Reddy, Courville, and Le~Roux}]{kazemnejad2024vineppo}
Amirhossein Kazemnejad, Milad Aghajohari, Eva Portelance, Alessandro Sordoni, Siva Reddy, Aaron Courville, and Nicolas Le~Roux. 2024.
\newblock Vineppo: Unlocking rl potential for llm reasoning through refined credit assignment.

\bibitem[{Li et~al.(2025{\natexlab{a}})Li, Elmahdy, Boyd, Wang, Garcia, Bhatia, Kass-Hout, Xiao, and Hong}]{li2025st}
Chenliang Li, Adel Elmahdy, Alex Boyd, Zhongruo Wang, Alfredo Garcia, Parminder Bhatia, Taha Kass-Hout, Cao Xiao, and Mingyi Hong. 2025{\natexlab{a}}.
\newblock St-ppo: Stabilized off-policy proximal policy optimization for multi-turn agents training.
\newblock \emph{arXiv preprint arXiv:2511.20718}.

\bibitem[{Li et~al.(2025{\natexlab{b}})Li, Wang, and Liu}]{lipipa}
Junbo Li, Zhangyang Wang, and Qiang Liu. 2025{\natexlab{b}}.
\newblock Pipa: Preference alignment as prior-informed statistical estimation.
\newblock In \emph{Forty-second International Conference on Machine Learning}.

\bibitem[{Li et~al.(2024)Li, Zhao, Wang, Wang, Zhou, Srivastava, Gokmen, Lee, Li, Zhang et~al.}]{li2024embodied}
Manling Li, Shiyu Zhao, Qineng Wang, Kangrui Wang, Yu~Zhou, Sanjana Srivastava, Cem Gokmen, Tony Lee, Erran~Li Li, Ruohan Zhang, and 1 others. 2024.
\newblock Embodied agent interface: Benchmarking llms for embodied decision making.
\newblock \emph{Advances in Neural Information Processing Systems}, 37:100428--100534.

\bibitem[{Li et~al.(2023)Li, Xu, Zhang, Lin, Yu, Sun, and Luo}]{li2023remax}
Ziniu Li, Tian Xu, Yushun Zhang, Zhihang Lin, Yang Yu, Ruoyu Sun, and Zhi-Quan Luo. 2023.
\newblock Remax: A simple, effective, and efficient reinforcement learning method for aligning large language models.
\newblock \emph{arXiv preprint arXiv:2310.10505}.

\bibitem[{Liu et~al.(2025{\natexlab{a}})Liu, Chen, Li, Qi, Pang, Du, Lee, and Lin}]{liu2025understanding}
Zichen Liu, Changyu Chen, Wenjun Li, Penghui Qi, Tianyu Pang, Chao Du, Wee~Sun Lee, and Min Lin. 2025{\natexlab{a}}.
\newblock Understanding r1-zero-like training: A critical perspective.
\newblock \emph{arXiv preprint arXiv:2503.20783}.

\bibitem[{Liu et~al.(2025{\natexlab{b}})Liu, Sims, Duan, Chen, Yu, Zhou, Xu, Xiong, Liu, Tan et~al.}]{liu2025gem}
Zichen Liu, Anya Sims, Keyu Duan, Changyu Chen, Simon Yu, Xiangxin Zhou, Haotian Xu, Shaopan Xiong, Bo~Liu, Chenmien Tan, and 1 others. 2025{\natexlab{b}}.
\newblock Gem: A gym for agentic llms.
\newblock \emph{arXiv preprint arXiv:2510.01051}.

\bibitem[{Murphy(2024)}]{murphy2024reinforcement}
Kevin Murphy. 2024.
\newblock Reinforcement learning: an overview.
\newblock \emph{arXiv preprint arXiv:2412.05265}.

\bibitem[{Putta et~al.(2024)Putta, Mills, Garg, Motwani, Finn, Garg, and Rafailov}]{putta2024agent}
Pranav Putta, Edmund Mills, Naman Garg, Sumeet Motwani, Chelsea Finn, Divyansh Garg, and Rafael Rafailov. 2024.
\newblock Agent q: Advanced reasoning and learning for autonomous ai agents.
\newblock \emph{arXiv preprint arXiv:2408.07199}.

\bibitem[{Qi et~al.(2024)Qi, Liu, Iong, Lai, Sun, Zhao, Yang, Yang, Sun, Yao et~al.}]{qi2024webrl}
Zehan Qi, Xiao Liu, Iat~Long Iong, Hanyu Lai, Xueqiao Sun, Wenyi Zhao, Yu~Yang, Xinyue Yang, Jiadai Sun, Shuntian Yao, and 1 others. 2024.
\newblock Webrl: Training llm web agents via self-evolving online curriculum reinforcement learning.
\newblock \emph{arXiv preprint arXiv:2411.02337}.

\bibitem[{Qin et~al.(2025)Qin, Ye, Fang, Wang, Liang, Tian, Zhang, Li, Li, Huang et~al.}]{qin2025ui}
Yujia Qin, Yining Ye, Junjie Fang, Haoming Wang, Shihao Liang, Shizuo Tian, Junda Zhang, Jiahao Li, Yunxin Li, Shijue Huang, and 1 others. 2025.
\newblock Ui-tars: Pioneering automated gui interaction with native agents.
\newblock \emph{arXiv preprint arXiv:2501.12326}.

\bibitem[{Rafailov et~al.(2023)Rafailov, Sharma, Mitchell, Manning, Ermon, and Finn}]{rafailov2023direct}
Rafael Rafailov, Archit Sharma, Eric Mitchell, Christopher~D Manning, Stefano Ermon, and Chelsea Finn. 2023.
\newblock Direct preference optimization: Your language model is secretly a reward model.
\newblock \emph{Advances in neural information processing systems}, 36:53728--53741.

\bibitem[{Schulman et~al.(2017)Schulman, Wolski, Dhariwal, Radford, and Klimov}]{schulman2017proximal}
John Schulman, Filip Wolski, Prafulla Dhariwal, Alec Radford, and Oleg Klimov. 2017.
\newblock Proximal policy optimization algorithms.
\newblock \emph{arXiv preprint arXiv:1707.06347}.

\bibitem[{Shao et~al.(2024)Shao, Wang, Zhu, Xu, Song, Bi, Zhang, Zhang, Li, Wu et~al.}]{shao2024deepseekmath}
Zhihong Shao, Peiyi Wang, Qihao Zhu, Runxin Xu, Junxiao Song, Xiao Bi, Haowei Zhang, Mingchuan Zhang, YK~Li, Yang Wu, and 1 others. 2024.
\newblock Deepseekmath: Pushing the limits of mathematical reasoning in open language models.
\newblock \emph{arXiv preprint arXiv:2402.03300}.

\bibitem[{Team(2024)}]{team2024qwen2}
Qwen Team. 2024.
\newblock Qwen2 technical report.
\newblock \emph{arXiv preprint arXiv:2407.10671}.

\bibitem[{Wang et~al.(2025{\natexlab{a}})Wang, Zou, Song, Feng, Fang, Lu, Liu, Luo, Liang, Huang et~al.}]{wang2025ui}
Haoming Wang, Haoyang Zou, Huatong Song, Jiazhan Feng, Junjie Fang, Junting Lu, Longxiang Liu, Qinyu Luo, Shihao Liang, Shijue Huang, and 1 others. 2025{\natexlab{a}}.
\newblock Ui-tars-2 technical report: Advancing gui agent with multi-turn reinforcement learning.
\newblock \emph{arXiv preprint arXiv:2509.02544}.

\bibitem[{Wang et~al.(2025{\natexlab{b}})Wang, Wang, Wang, Zhang, Li, Yang, Jin, Yu, Nguyen, Liu et~al.}]{wang2025ragen}
Zihan Wang, Kangrui Wang, Qineng Wang, Pingyue Zhang, Linjie Li, Zhengyuan Yang, Xing Jin, Kefan Yu, Minh~Nhat Nguyen, Licheng Liu, and 1 others. 2025{\natexlab{b}}.
\newblock Ragen: Understanding self-evolution in llm agents via multi-turn reinforcement learning.
\newblock \emph{arXiv preprint arXiv:2504.20073}.

\bibitem[{Wei et~al.(2025)Wei, Yao, Liu, Zhang, Lu, Qiu, Yu, Xu, Zhang, Yin et~al.}]{wei2025webagent}
Zhepei Wei, Wenlin Yao, Yao Liu, Weizhi Zhang, Qin Lu, Liang Qiu, Changlong Yu, Puyang Xu, Chao Zhang, Bing Yin, and 1 others. 2025.
\newblock Webagent-r1: Training web agents via end-to-end multi-turn reinforcement learning.
\newblock \emph{arXiv preprint arXiv:2505.16421}.

\bibitem[{Williams(1992)}]{williams1992simple}
Ronald~J Williams. 1992.
\newblock Simple statistical gradient-following algorithms for connectionist reinforcement learning.
\newblock \emph{Machine learning}, 8(3):229--256.

\bibitem[{Xiong et~al.(2025)Xiong, Yao, Xu, Pang, Wang, Sahoo, Li, Jiang, Zhang, Xiong et~al.}]{xiong2025minimalist}
Wei Xiong, Jiarui Yao, Yuhui Xu, Bo~Pang, Lei Wang, Doyen Sahoo, Junnan Li, Nan Jiang, Tong Zhang, Caiming Xiong, and 1 others. 2025.
\newblock A minimalist approach to llm reasoning: from rejection sampling to reinforce.
\newblock \emph{arXiv preprint arXiv:2504.11343}.

\bibitem[{Yang et~al.(2025)Yang, Li, Yang, Zhang, Hui, Zheng, Yu, Gao, Huang, Lv et~al.}]{yang2025qwen3}
An~Yang, Anfeng Li, Baosong Yang, Beichen Zhang, Binyuan Hui, Bo~Zheng, Bowen Yu, Chang Gao, Chengen Huang, Chenxu Lv, and 1 others. 2025.
\newblock Qwen3 technical report.
\newblock \emph{arXiv preprint arXiv:2505.09388}.

\bibitem[{Yao et~al.(2022)Yao, Chen, Yang, and Narasimhan}]{yao2022webshop}
Shunyu Yao, Howard Chen, John Yang, and Karthik Narasimhan. 2022.
\newblock Webshop: Towards scalable real-world web interaction with grounded language agents.
\newblock \emph{Advances in Neural Information Processing Systems}, 35:20744--20757.

\bibitem[{Yao et~al.(2023)Yao, Zhao, Yu, Du, Shafran, Narasimhan, and Cao}]{yao2023react}
Shunyu Yao, Jeffrey Zhao, Dian Yu, Nan Du, Izhak Shafran, Karthik Narasimhan, and Yuan Cao. 2023.
\newblock React: Synergizing reasoning and acting in language models.
\newblock In \emph{International Conference on Learning Representations (ICLR)}.

\bibitem[{Yin et~al.(2025)Yin, Yang, Xie, Yang, Sun, Awadalla, Chen, and Zhou}]{yin2025segmenting}
Yueqin Yin, Shentao Yang, Yujia Xie, Ziyi Yang, Yuting Sun, Hany Awadalla, Weizhu Chen, and Mingyuan Zhou. 2025.
\newblock Segmenting text and learning their rewards for improved rlhf in language model.
\newblock \emph{arXiv preprint arXiv:2501.02790}.

\bibitem[{Yu et~al.(2025)Yu, Zhang, Zhu, Yuan, Zuo, Yue, Dai, Fan, Liu, Liu et~al.}]{yu2025dapo}
Qiying Yu, Zheng Zhang, Ruofei Zhu, Yufeng Yuan, Xiaochen Zuo, Yu~Yue, Weinan Dai, Tiantian Fan, Gaohong Liu, Lingjun Liu, and 1 others. 2025.
\newblock Dapo: An open-source llm reinforcement learning system at scale.
\newblock \emph{arXiv preprint arXiv:2503.14476}.

\bibitem[{Yuan et~al.(2025)Yuan, Yue, Zhu, Fan, and Yan}]{yuan2025s}
Yufeng Yuan, Yu~Yue, Ruofei Zhu, Tiantian Fan, and Lin Yan. 2025.
\newblock What's behind ppo's collapse in long-cot? value optimization holds the secret.
\newblock \emph{arXiv preprint arXiv:2503.01491}.

\bibitem[{Yue et~al.(2025)Yue, Yuan, Yu, Zuo, Zhu, Xu, Chen, Wang, Fan, Du et~al.}]{yue2025vapo}
Yu~Yue, Yufeng Yuan, Qiying Yu, Xiaochen Zuo, Ruofei Zhu, Wenyuan Xu, Jiaze Chen, Chengyi Wang, TianTian Fan, Zhengyin Du, and 1 others. 2025.
\newblock Vapo: Efficient and reliable reinforcement learning for advanced reasoning tasks.
\newblock \emph{arXiv preprint arXiv:2504.05118}.

\bibitem[{Zeng et~al.(2025)Zeng, Wei, Brown, Frunza, Nevmyvaka, and Hong}]{zeng2025reinforcing}
Siliang Zeng, Quan Wei, William Brown, Oana Frunza, Yuriy Nevmyvaka, and Mingyi Hong. 2025.
\newblock Reinforcing multi-turn reasoning in llm agents via turn-level credit assignment.
\newblock \emph{arXiv preprint arXiv:2505.11821}.

\bibitem[{Zheng et~al.(2025)Zheng, Liu, Li, Chen, Yu, Gao, Dang, Liu, Men, Yang et~al.}]{zheng2025group}
Chujie Zheng, Shixuan Liu, Mingze Li, Xiong-Hui Chen, Bowen Yu, Chang Gao, Kai Dang, Yuqiong Liu, Rui Men, An~Yang, and 1 others. 2025.
\newblock Group sequence policy optimization.
\newblock \emph{arXiv preprint arXiv:2507.18071}.

\bibitem[{Zhou et~al.(2024)Zhou, Zanette, Pan, Levine, and Kumar}]{zhou2024archer}
Yifei Zhou, Andrea Zanette, Jiayi Pan, Sergey Levine, and Aviral Kumar. 2024.
\newblock Archer: Training language model agents via hierarchical multi-turn rl.
\newblock \emph{arXiv preprint arXiv:2402.19446}.

\end{thebibliography}

\newpage
\appendix
\section{Training details}
We report the full training hyperparameters used in our main experiments. Following the notation in Section~\ref{sec:ppo exp}, we denote \((B_R, G, B_M, E)\) as the total number of rollout samples per iteration, the group size per distinct query, the minibatch size, and the number of epochs over each rollout, respectively. For GRPO, we use a learning rate of \(1\times10^{-6}\), a KL coefficient of \(0.001\), and set \((B_R, G, B_M, E)=(256,16,64,1)\) for WebShop and \((512,16,128,1)\) for Sokoban. For PPO, we use an actor learning rate of \(1\times10^{-6}\), a critic learning rate of \(1\times10^{-5}\), with \((B_R, G, B_M, E)=(256,1,64,1)\) for WebShop and \((512,16,128,1)\) for Sokoban. For both GRPO and PPO, the clip ratio is fixed at \(0.2\), with 100 training steps for Qwen2.5\text{-}7B and 200 steps for Qwen2.5\text{-}3B and Qwen3\text{-}1.7B.

\section{Trajectory examples}
\label{sec: example}
We present example trajectories from Qwen2.5 and Qwen3 on WebShop. These examples clearly reveal distinct reasoning patterns between the two models. In particular, Qwen3 exhibits much longer reasoning chains, which substantially increases the difficulty of training.

\newtcolorbox{trajectorybox}[1][]{%
  enhanced,
  colback=gray!8,
  colframe=black,
  boxrule=0.6pt,
  % rounded corners=2pt,
  breakable,
  left=6pt,right=6pt,top=6pt,bottom=6pt,
  before skip=10pt,after skip=10pt,
  title=#1
}

{\small
\begin{trajectorybox}[Qwen2.5]
\textbf{system}\\
You're a helpful assistant.\\[6pt]

\textbf{user}\\
You are browsing an online shop. Based on the instruction, find the product that close to the production description. You need to read the website and decide what action to take next until buying a product. Available actions depends on the page: in the search page you can search keywords, in the search result page you can click an item url or click[next] to navigate to next page, in the product page you can click[description] or click[features] to see the details, click[blue] or click[x-large] to choose size and colors, click[buy now] when you decided to buy the product, click[back to search] to return to search page. You should only choose action from the available actions list.  Example process: I need a gingko light and 20x20 pillow cover that is hand painted. First search[gingko light 20x20 pillow cover hand painted], answer format: \verb|<answer>search[blanket with fleece throw]</answer>|. Valid answer is search[\textless{}keywords\textgreater{}] or click[\textless{}clickable\textgreater{}].\\[6pt]

\textbf{Turn 1:}\\
\textbf{State:}\\
Instruction: Find me machine wash, wash cold women's fashion hoodies \& sweatshirts for dry clean, tumble dry with color: vintage camo, and size: large, and price lower than 60.00 dollars. We must buy a product within 10 actions. It doesn't have to match perfectly with description. Search term should not include details like size, color. Never search for more than 2 times. Do not be too strict about the description, it's more important to buy one that is close enough within action limit. Prioritize click a product in the current page over going to next page. Almost never click[next] for more than 2 times. Almost never click[\textless{} prev] unless you are sure the product is on one of the previous pages. If you have less than 3 actions left, just buy the first product you see in the current page. If an matching option exists, make sure to click[size] then click[color], one at a time, before click[buy now], but don't have to if only 1 action left, in that case you just click[buy now]. Never click description. You must choose from these actions: search[\textless{}content\textgreater{}]. You have 9 actions left. Always output: \verb|<think>...</think><answer>...</answer>| with no extra text. Strictly follow this format. Max response length: 300 words (tokens).\\[2pt]

\textbf{assistant}\\
\texttt{<think>Based on the instruction, I need to find machine wash, wash cold women's fashion hoodies \& sweatshirts for dry clean, tumble dry with color: vintage camo, and size: large, with a price lower than 60.00 dollars. I will start by searching for these keywords to find a relevant product.</think><answer>search[holiday hoodie vintage camo large dry clean]</answer>}\\[6pt]

% \textbf{user}\\
\textbf{Reward:} 0.0\\[8pt]

\textbf{Turn 2:}\\
\textbf{State:}\\
Instruction: [SEP] Find me machine wash, wash cold women's fashion hoodies \& sweatshirts for dry clean, tumble dry with color: vintage camo, and size: large, and price lower than 60.00 dollars [SEP] Back to Search [SEP] Page 1 (Total results: 50) [SEP] next [SEP] B01HQTWL6S [SEP] Expression Tees Pug Life Funny Thug Life Unisex Adult Hoodie [SEP] \$36.99 to \$48.99 [SEP] B09QQP3356 [SEP] HAUKLIE Men's Sports Waffle Ribbed Polo Shirts Summer Short Sleeve Cotton Muscle Quarter-Zip Henley T-Shirt Tunics Tops [SEP] \$10.99 [SEP] B07WX8FPGG [SEP] Matching Couple Hoodies Set Funny Hubby Wifey Hoodies Pullover Sweater Honeymoon [SEP] \$23.99 to \$37.99 [SEP] B09HGFV91W [SEP] ZHDD Tops for Mens, Men's Knight Punk Vintage Style Long Sleeve Asymmetric Hem Cowl Neck Teen Boys Gothic Pullover [SEP] \$4.98 to \$7.99 [SEP] B09M63B87V [SEP] YALFJV Women Long Sleeve Crew Neck Side Button T Shirts Tunic Dress Loose Asymmetric Hem Tunic Pullover to Wear with Leggings [SEP] \$10.71 to \$18.34 [SEP] B00O30JLDK [SEP] Legendary Whitetails Men's Non-Typical Long Sleeve T-Shirt [SEP] \$10.52 to \$40.5 [SEP] B07Q87P8DQ [SEP] Permo Vintage Rustic Industrial 3-Lights Kitchen Island Chandelier Triple 3 Heads Pendant Hanging Ceiling Lighting Fixture with Oval Cone Clear Glass Shade (Antique) [SEP] \$94.99 [SEP] B09Q8RD8YN [SEP] Bungo Stray Anime Dogs Anime Character, Long Sleeve, Sweatshirt, Hoodie, T shirt [SEP] \$19.99 [SEP] B09QQJJ3KM [SEP] One Lucky Teacher St Patrick Day Shamrock Tee Teachers Custom Personalized Unisex T-Shirts Long Sleeve Hoodie Sweatshirt Gifts [SEP] \$100.0 [SEP] B09S3BN15C [SEP] Mens Linen Shirt,Men's Striped Shirts Casual Short Sleeve Button Down Shirts Regular Fit Hawaiian Shirts Beach Tees Tops [SEP] \$3.78 to \$11.38. We must buy a product within 10 actions. It doesn't have to match perfectly with description. Search term should not include details like size, color. Never search for more than 2 times. Do not be too strict about the description, it's more important to buy one that is close enough within action limit. Prioritize click a product in the current page over going to next page. Almost never click[next] for more than 2 times. Almost never click[\textless{} prev] unless you are sure the product is on one of the previous pages. If you have less than 3 actions left, just buy the first product you see in the current page. If an matching option exists, make sure to click[size] then click[color], one at a time, before click[buy now], but don't have to if only 1 action left, in that case you just click[buy now]. Never click description. You must choose from these actions: click[back to search], click[next], click[b01hqtwl6s], click[b09qqp3356], click[b07wx8fpgg], click[b09hgfv91w], click[b09m63b87v], click[b00o30jldk], click[b07q87p8dq], click[b09q8rd8yn], click[b09qqjj3km], click[b09s3bn15c]. You have 8 actions left. Always output: \verb|<think>...</think><answer>...</answer>| with no extra text. Strictly follow this format. Max response length: 300 words (tokens).\\[2pt]

\textbf{assistant}\\
\texttt{<think>Based on the search results, I need to find a women's fashion hoodie with a vintage camo color and size large, priced lower than 60.00 dollars. The product B01HQTWL6S appears to be a holiday hoodie with vintage camo color and a price range that fits the criteria. I will click on this product to proceed to the next steps.</think><answer>click[b01hqtwl6s]</answer>}\\[6pt]

% \textbf{user}\\
\textbf{Reward:} 0.0\\[8pt]

\textbf{Turn 3:}\\
\textbf{State:}\\
Instruction: [SEP] Find me machine wash, wash cold women's fashion hoodies \& sweatshirts for dry clean, tumble dry with color: vintage camo, and size: large, and price lower than 60.00 dollars [SEP] Back to Search [SEP] \textless{} Prev [SEP] size [SEP] small [SEP] medium [SEP] large [SEP] x-large [SEP] xx-large [SEP] 3x-large [SEP] 4x-large [SEP] color [SEP] black [SEP] charcoal grey [SEP] forest green [SEP] heather grey [SEP] kelly green [SEP] maroon [SEP] navy blue [SEP] red [SEP] royal blue [SEP] white [SEP] cotton candy [SEP] sand camo [SEP] vintage camo [SEP] Expression Tees Pug Life Funny Thug Life Unisex Adult Hoodie [SEP] Price: \$36.99 to \$48.99 [SEP] Rating: N.A. [SEP] Description [SEP] Features [SEP] Reviews [SEP] Buy Now. We must buy a product within 10 actions. ... You must choose from these actions: click[back to search], click[\textless{} prev], click[description], click[features], click[reviews], click[buy now], click[small], click[medium], click[large], click[x-large], click[xx-large], click[3x-large], click[4x-large], click[black], click[charcoal grey], click[forest green], click[heather grey], click[kelly green], click[maroon], click[navy blue], click[red], click[royal blue], click[white], click[cotton candy], click[sand camo], click[vintage camo]. You have 7 actions left.\\[2pt]

\textbf{assistant}\\
\texttt{|<think>Based on the current page, I need to find the large size and vintage camo color hoodie. The product B01HQTWL6S is the correct one, and I need to prioritize clicking the size and color options before proceeding to buy.</think><answer>click[large]</answer>}\\[6pt]

% \textbf{user}\\
\textbf{Reward:} 0.0\\[8pt]

\textbf{Turn 4:}\\
\textbf{State:} (same as Turn 3)\\[2pt]
\textbf{assistant}\\
\texttt{<think>Based on the current page, I have clicked on the large size. Now, I need to prioritize clicking the color option to ensure I find the correct hoodie with the vintage camo color. The color options include various choices, and vintage camo is one of them.</think><answer>click[vintage camo]</answer>}\\[6pt]

% \textbf{user}\\
\textbf{Reward:} 0.0\\[8pt]

\textbf{Turn 5:}\\
\textbf{State:} (same as Turn 3)\\[2pt]
\textbf{assistant}\\
\texttt{<think>Based on the current page, I have clicked on the large size and vintage camo color. Now, I need to proceed to the buy now button to complete the purchase. I will click on the buy now button to finalize the purchase.</think><answer>click[buy now]</answer>}\\[6pt]

\textbf{Reward:} 0.8
\end{trajectorybox}
}

{\small
\begin{trajectorybox}[Qwen3]
\textbf{system}\\
You're a helpful assistant.\\[6pt]

\textbf{user}\\
You are browsing an online shop. Based on the instruction, find the product that close to the production description. You need to read the website and decide what action to take next until buying a product. Available actions depends on the page: in the search page you can search keywords, in the search result page you can click an item url or click[next] to navigate to next page, in the product page you can click[description] or click[features] to see the details, click[blue] or click[x-large] to choose size and colors, click[buy now] when you decided to buy the product, click[back to search] to return to search page. You should only choose action from the available actions list.  Example process: I need a gingko light and 20x20 pillow cover that is hand painted. First search[gingko light 20x20 pillow cover hand painted], answer format: \texttt{<answer>search[blanket with fleece throw]</answer>}. Valid answer is search[\textless{}keywords\textgreater{}] or click[\textless{}clickable\textgreater{}].\\[6pt]

\textbf{Turn 1:}\\
\textbf{State:}\\
Instruction: Find me loose fit, day comfort, hand wash women's tops, tees \& blouses with short sleeve, polyester spandex for teen girls with color: b15-red, and size: 3x-large, and price lower than 40.00 dollars.We must buy a product within 10 actions. It doesn't have to match perfectly with description.\\
Search term should not include details like size, color.\\
Never search for more than 2 times.\\
Do not be too strict about the description, it's more important to buy one that is close enough within action limit.\\
Prioritize click a product in the current page over going to next page.\\
Almost never click[next] for more than 2 times.Almost never click[\textless{} prev] unless you are sure the product is on one of the previous pages.\\
If you have less than 3 actions left, just buy the first product you see in the current page.\\
If an matching option exists, make sure to click[size] then click[color], one at a time, before click[buy now], but don't have to if only 1 action left, in that case you just click[buy now]. Never click description.\\
You must choose from these actions:search[\textless{}content\textgreater{}].\\
You have 9 actions left. Always output: \texttt{<think> [Your thoughts] </think> <answer> [your answer] </answer>} with no extra text. Strictly follow this format. Max response length: 300 words (tokens).\\[2pt]

\textbf{assistant}\\
\texttt{<think>
Okay, let's see. The user is looking for loose fit, day comfort, hand wash women's tops, tees \& blouses with short sleeve, polyester spandex for teen girls in b15-red, size 3x-large, under \$40. They need to buy within 10 actions.
First, I need to search for the right keywords. The search term should not include size or color details, just the main features. So maybe "loose fit day comfort short sleeve polyester spandex teen girls hand wash" could work. That way, the search results can show products that match the description even if the exact size and color aren't specified.
Next, after searching, the user would see the results. They need to click on a product. Since the example shows clicking on a product URL, I'll assume the first result is the one they want. Then, they might need to select the color and size. But since the user can only click once per action, maybe they can click buy now immediately if the product meets the criteria. However, the instruction says to click[size] then[color] if needed, but if only one action left, just buy. So if the first product meets all the criteria, they can buy it right away. Otherwise, move to the next page. But the user said not to search more than twice, so maybe the first search is enough. Let's go with the first search and then buy the first product found.
</think>
<answer>search[loose fit day comfort short sleeve polyester spandex teen girls hand wash]</answer>}\\[6pt]

% \textbf{user}\\
\textbf{Reward:} 0.0\\[8pt]

\textbf{Turn 2:}\\
\textbf{State:}\\
Instruction: [SEP] Find me loose fit, day comfort, hand wash women's tops, tees \& blouses with short sleeve, polyester spandex for teen girls with color: b15-red, and size: 3x-large, and price lower than 40.00 dollars [SEP] Back to Search [SEP] Page 1 (Total results: 50) [SEP] next [SEP] B09QQP3356 [SEP] HAUKLIE Men's Sports Waffle Ribbed Polo Shirts Summer Short Sleeve Cotton Muscle Quarter-Zip Henley T-Shirt Tunics Tops [SEP] \$10.99 [SEP] B09M63B87V [SEP] YALFJV Women Long Sleeve Crew Neck Side Button T Shirts Tunic Dress Loose Asymmetric Hem Tunic Pullover to Wear with Leggings [SEP] \$10.71 to \$18.34 [SEP] B099231V35 [SEP] Mens Casual Cargo Pants Hi Vis Viz Reflective Overalls High Visibility Safe Work Pants Outdoor Hiking Trousers Big and Tall [SEP] \$16.79 [SEP] B09QCVCYVY [SEP] Burband Womens High Waist Bike Shorts Tummy Control Workout Yoga Pants Running Exercise Running Shorts with Side Pockets [SEP] \$6.82 [SEP] B09ND8P2QR [SEP] InterestPrint Deer and Stars Men's Loungewear Pajama Sets - Long Sleeve Tee and Jogger Pant [SEP] \$43.59 [SEP] B09ND9DP7J [SEP] InterestPrint Gold Horse Pattern Men's 2-Piece Sleepwear Set, Long Sleeve Shirt with Pants Loungewear [SEP] \$43.59 [SEP] B09KP78G37 [SEP] Women Faux Fur Lined Jacket Coat Winter Warm Thick Fleece Outwear Trench Zipper Plus Size Long Sleeve Plush Overcoat [SEP] \$47.41 to \$59.07 [SEP] B09Q5ZHRVM [SEP] Yinimo Mens Gym Shorts Fashionable Loose Beach Drawstring Turkey Print Beach Cool Shorts [SEP] \$19.19 [SEP] B09PBPZ24Z [SEP] myhehthw Women's High Waisted Jeans for Women Distressed Ripped Jeans Slim Fit Butt Lifting Skinny Stretch Jeans Trousers [SEP] \$22.99 to \$25.99 [SEP] B09S3BN15C [SEP] Mens Linen Shirt,Men's Striped Shirts Casual Short Sleeve Button Down Shirts Regular Fit Hawaiian Shirts Beach Tees Tops [SEP] \$3.78 to \$11.38.We must buy a product within 10 actions. It doesn't have to match perfectly with description.\\
Search term should not include details like size, color.\\
Never search for more than 2 times.\\
Do not be too strict about the description, it's more important to buy one that is close enough within action limit.\\
Prioritize click a product in the current page over going to next page.\\
Almost never click[next] for more than 2 times.Almost never click[\textless{} prev] unless you are sure the product is on one of the previous pages.\\
If you have less than 3 actions left, just buy the first product you see in the current page.\\
If an matching option exists, make sure to click[size] then click[color], one at a time, before click[buy now], but don't have to if only 1 action left, in that case you just click[buy now]. Never click description.\\
You must choose from these actions:click[back to search], click[next], click[b09qqp3356], click[b09m63b87v], click[b099231v35], click[b09qcvcyvy], click[b09nd8p2qr], click[b09nd9dp7j], click[b09kp78g37], click[b09q5zhrvm], click[b09pbpz24z], click[b09s3bn15c].\\
You have 8 actions left. Always output: \texttt{<think> [Your thoughts] </think> <answer> [your answer] </answer>} with no extra text. Strictly follow this format. Max response length: 300 words (tokens).\\[2pt]

\textbf{assistant}\\
\texttt{<think>
Okay, let's see. The user is looking for a specific product: loose fit, day comfort, short sleeve, polyester spandex, teen girls, red color, 3x-large, under \$40. They have 8 actions left.
In Turn 1, they searched for "loose fit day comfort short sleeve polyester spandex teen girls hand wash" and got several results. The first result is a polo shirt priced at \$10.99. The user needs to check if this meets the criteria. The color is not specified in the search term, but the product's color is mentioned as "b15-red" in the description. Since the search term didn't include color details, it's possible this could match. The size is 3x-large, which is part of the description, but the search term didn't mention size. The price is under \$40, which fits.
Now, the user has 8 actions left. They can click on the first product (B09QQP3356) to view it. Since the instruction says to click a product in the current page, they should click on the first result. Then, they can check if it meets all the criteria. If it does, they can proceed to buy it. If not, they might need to move to the next page, but the user can only click[next] twice. Since they have 8 actions left, clicking on the first product and then buying it would be efficient. The user needs to ensure they don't search again, so they should proceed to buy the first product found.
</think>
<answer>click[b09qqp3356]</answer>}\\[6pt]

% \textbf{user}\\
\textbf{Reward:} 0.0\\[8pt]

\textbf{Turn 3:}\\
\textbf{State:}\\
Instruction: [SEP] Find me loose fit, day comfort, hand wash women's tops, tees \& blouses with short sleeve, polyester spandex for teen girls with color: b15-red, and size: 3x-large, and price lower than 40.00 dollars [SEP] Back to Search [SEP] \textless{} Prev [SEP] color [SEP] black [SEP] gray [SEP] light blue [SEP] size [SEP] small [SEP] medium [SEP] large [SEP] x-large [SEP] xx-large [SEP] 3x-large [SEP] HAUKLIE Men's Sports Waffle Ribbed Polo Shirts Summer Short Sleeve Cotton Muscle Quarter-Zip Henley T-Shirt Tunics Tops [SEP] Price: \$10.99 [SEP] Rating: N.A. [SEP] Description [SEP] Features [SEP] Reviews [SEP] Buy Now.We must buy a product within 10 actions. It doesn't have to match perfectly with description.\\
Search term should not include details like size, color.\\
Never search for more than 2 times.\\
Do not be too strict about the description, it's more important to buy one that is close enough within action limit.\\
Prioritize click a product in the current page over going to next page.\\
Almost never click[next] for more than 2 times.Almost never click[\textless{} prev] unless you are sure the product is on one of the previous pages.\\
If you have less than 3 actions left, just buy the first product you see in the current page.\\
If an matching option exists, make sure to click[size] then click[color], one at a time, before click[buy now], but don't have to if only 1 action left, in that case you just click[buy now]. Never click description.\\
You must choose from these actions:click[back to search], click[\textless{} prev], click[description], click[features], click[reviews], click[buy now], click[black], click[gray], click[light blue], click[small], click[medium], click[large], click[x-large], click[xx-large], click[3x-large].\\
You have 7 actions left. Always output: \texttt{<think> [Your thoughts] </think> <answer> [your answer] </answer>} with no extra text. Strictly follow this format. Max response length: 300 words (tokens).\\[2pt]

\textbf{assistant}\\
\texttt{<think>
Okay, let's see. The user is trying to find a specific product: loose fit, day comfort, short sleeve, polyester spandex, teen girls, red color, 3x-large, under \$40. They have 7 actions left.
In Turn 2, they clicked on the first product (B09QQP3356) and now they're on the product page. The next step is to check if the color and size match. The product's color is not specified in the search term, but the user wants red (b15-red). The available colors are black, gray, light blue. Since red isn't listed, maybe the product is a different color. But the user might accept it as close. They need to select the color and size. The size is already 3x-large, so they can click that. Then, they can proceed to buy.
Since they have 7 actions left, they can click on the color options. If the product doesn't have red, they might need to move to the next page, but the user can only click[next] twice. Since they have 7 actions left, clicking on the color and size would take 2 actions, then buy now. But the user might need to check if the product meets all criteria. If it doesn't, they might have to search again, but the instruction says not to search more than twice. So, they should proceed to buy the first product they see, assuming it's close enough.
</think>
<answer>click[3x-large]</answer>}\\[6pt]

% \textbf{user}\\
\textbf{Reward:} 0.0\\[8pt]

\textbf{Turn 4:}\\
\textbf{State:}\\
Instruction: [SEP] Find me loose fit, day comfort, hand wash women's tops, tees \& blouses with short sleeve, polyester spandex for teen girls with color: b15-red, and size: 3x-large, and price lower than 40.00 dollars [SEP] Back to Search [SEP] \textless{} Prev [SEP] color [SEP] black [SEP] gray [SEP] light blue [SEP] size [SEP] small [SEP] medium [SEP] large [SEP] x-large [SEP] xx-large [SEP] 3x-large [SEP] HAUKLIE Men's Sports Waffle Ribbed Polo Shirts Summer Short Sleeve Cotton Muscle Quarter-Zip Henley T-Shirt Tunics Tops [SEP] Price: \$10.99 [SEP] Rating: N.A. [SEP] Description [SEP] Features [SEP] Reviews [SEP] Buy Now.We must buy a product within 10 actions. It doesn't have to match perfectly with description.\\
Search term should not include details like size, color.\\
Never search for more than 2 times.\\
Do not be too strict about the description, it's more important to buy one that is close enough within action limit.\\
Prioritize click a product in the current page over going to next page.\\
Almost never click[next] for more than 2 times.Almost never click[\textless{} prev] unless you are sure the product is on one of the previous pages.\\
If you have less than 3 actions left, just buy the first product you see in the current page.\\
If an matching option exists, make sure to click[size] then click[color], one at a time, before click[buy now], but don't have to if only 1 action left, in that case you just click[buy now]. Never click description.\\
You must choose from these actions:click[back to search], click[\textless{} prev], click[description], click[features], click[reviews], click[buy now], click[black], click[gray], click[light blue], click[small], click[medium], click[large], click[x-large], click[xx-large], click[3x-large].\\
You have 6 actions left. Always output: \texttt{<think> [Your thoughts] </think> <answer> [your answer] </answer>} with no extra text. Strictly follow this format. Max response length: 300 words (tokens).\\[2pt]

\textbf{assistant}\\
\texttt{<think>
Okay, let's see. The user is trying to find a specific product: loose fit, day comfort, short sleeve, polyester spandex, teen girls, red color, 3x-large, under \$40. They have 6 actions left.
In Turn 3, they clicked on the size 3x-large. Now, they need to check the color. The product's color isn't specified, but the user wants red (b15-red). The available colors are black, gray, light blue. Since red isn't listed, maybe the product is a different color. But the user might accept it as close. They can click on the color options. Since they have 6 actions left, they can try the colors. If the product doesn't have red, they might need to move to the next page, but the user can only click[next] twice. Since they have 6 actions left, they can try the colors and then buy. But the instruction says to buy within 10 actions. So, they should proceed to buy after checking the color. However, the user might need to click on the color first. Wait, in Turn 3, they clicked on 3x-large, and now they're on the product page. The next step is to select the color. The available colors are black, gray, light blue. The user can click on these. If the product doesn't have red, they might need to move to the next page, but the user can only click[next] twice. Since they have 6 actions left, they can click on the color options. Then, after selecting the color, they can buy. But the user might need to check if the color matches. If it doesn't, they might have to search again, but the instruction says not to search more than twice. So, they should proceed to buy after selecting the color. The final action would be to click buy now once the color is selected. But the user has 6 actions left, so they can click on the color, then buy. Therefore, the answer is to click on the color and then buy.
</think>
<answer>click[black]</answer>}\\[6pt]

% \textbf{user}\\
\textbf{Reward:} 0.0\\[8pt]

\textbf{Turn 5:}\\
\textbf{State:}\\
Instruction: [SEP] Find me loose fit, day comfort, hand wash women's tops, tees \& blouses with short sleeve, polyester spandex for teen girls with color: b15-red, and size: 3x-large, and price lower than 40.00 dollars [SEP] Back to Search [SEP] \textless{} Prev [SEP] color [SEP] black [SEP] gray [SEP] light blue [SEP] size [SEP] small [SEP] medium [SEP] large [SEP] x-large [SEP] xx-large [SEP] 3x-large [SEP] HAUKLIE Men's Sports Waffle Ribbed Polo Shirts Summer Short Sleeve Cotton Muscle Quarter-Zip Henley T-Shirt Tunics Tops [SEP] Price: \$10.99 [SEP] Rating: N.A. [SEP] Description [SEP] Features [SEP] Reviews [SEP] Buy Now.We must buy a product within 10 actions. It doesn't have to match perfectly with description.\\
Search term should not include details like size, color.\\
Never search for more than 2 times.\\
Do not be too strict about the description, it's more important to buy one that is close enough within action limit.\\
Prioritize click a product in the current page over going to next page.\\
Almost never click[next] for more than 2 times.Almost never click[\textless{} prev] unless you are sure the product is on one of the previous pages.\\
If you have less than 3 actions left, just buy the first product you see in the current page.\\
If an matching option exists, make sure to click[size] then click[color], one at a time, before click[buy now], but don't have to if only 1 action left, in that case you just click[buy now]. Never click description.\\
You must choose from these actions:click[back to search], click[\textless{} prev], click[description], click[features], click[reviews], click[buy now], click[black], click[gray], click[light blue], click[small], click[medium], click[large], click[x-large], click[xx-large], click[3x-large].\\
You have 5 actions left. Always output: \texttt{<think> [Your thoughts] </think> <answer> [your answer] </answer>} with no extra text. Strictly follow this format. Max response length: 300 words (tokens).\\[2pt]

\textbf{assistant}\\
\texttt{<think>
Okay, let's see. The user is trying to find a specific product: loose fit, day comfort, short sleeve, polyester spandex, teen girls, red color, 3x-large, under \$40. They have 5 actions left.
In Turn 4, they clicked on the color black. Now, they need to check if the product meets all criteria. The color is black, but the user wanted red (b15-red). Since the search term didn't include color details, it's possible the product is a different color. But the user might accept it as close. They have 5 actions left. The next step is to buy the product. Since they have 5 actions left, they can click buy now immediately after selecting the color. However, the instruction says to click[size] then[color] before buy, but if only 1 action left, just buy. Since they have 5 actions left, they can proceed to buy now. So the answer is to click buy now.
</think>
<answer>click[buy now]</answer>}\\[6pt]

\textbf{Reward:} 0.6

\end{trajectorybox}
}

\end{document}